\pdfoutput=1
\documentclass[runningheads]{llncs}

\usepackage{microtype}
\usepackage{enumitem}
\usepackage{graphicx}
\usepackage{subfigure}
\usepackage{booktabs} 
\usepackage{tabularx}
\usepackage{ragged2e}
\usepackage[table]{xcolor}

\usepackage{hyperref}

\usepackage[abbreviations]{foreign}  
\usepackage{xcolor}
\usepackage[T1]{fontenc} 
\usepackage{url}

\usepackage{breakurl}

\usepackage[]{hyperref}
\hypersetup{
  colorlinks,
  linkcolor={green!50!black},
  citecolor={red!70!black},
  urlcolor={blue!70!black}
}

\usepackage[acronym]{glossaries}

\usepackage{listings}
\newcommand\YAMLkeystyle{\color{black}\scriptsize\scriptsize\ttfamily\bfseries}
\newcommand\YAMLvaluestyle{\color{black}\scriptsize\scriptsize\ttfamily}
\newcommand\YAMLcommentstyle{\color{gray!70!black}\ttfamily}
\makeatletter
\newcommand\language@yaml{yaml}
\expandafter\expandafter\expandafter\lstdefinelanguage
\expandafter{\language@yaml}
{
	keywords={true,false,null,y,n},
	basicstyle=\linespread{0.8}\color{black}\ttfamily,  
	keywordstyle=\color{darkgray}\bfseries,
	sensitive=false,
	comment=[l]{\#},
	morecomment=[s]{/*}{*/},
	commentstyle=\YAMLcommentstyle,
	stringstyle=\YAMLvaluestyle,
	moredelim=[l][\color{orange}]{\&},
	moredelim=[l][\color{magenta}]{*},
	moredelim=**[il][\YAMLvaluestyle]{},  
	morestring=[b]',
	morestring=[b]",
	literate = {---}{{\ProcessThreeDashes}}3
	{>}{{\textcolor{red!70!black}\textgreater}}1
	{|}{{\textcolor{red!70!black}\textbar}}1
	{-}{{\mdseries-}}1,
}

\lst@AddToHook{EveryLine}{\ifx\lst@language\language@yaml\YAMLkeystyle\fi}
\makeatother

\lstdefinelanguage{Ini}
{
	basicstyle=\fontsize{10}{10}\ttfamily,
	morecomment=[s][\color{blue}]{[}{]},
	morecomment=[l]{\#},
	morecomment=[l]{;},
	commentstyle=\color{gray},
	morekeywords={},
	otherkeywords={=},
	keywordstyle={\color{gray}}
}
\usepackage{MnSymbol}

\newacronym{sgx}{SGX}{Intel software guard extensions}
\newacronym{txt}{TXT}{Intel trusted execution technology}
\newacronym{tee}{TEE}{trusted execution environment}
\newacronym{kms}{KMS}{key management system}
\newacronym{epc}{EPC}{enclave page cache}
\newacronym{tls}{TLS}{transport layer security}
\newacronym{os}{OS}{operating system}
\newacronym{drtm}{DRTM}{dynamic root of trust for measurements}
\newacronym{srtm}{SRTM}{static root of trust for measurements}
\newacronym{crtm}{CRTM}{core root of trust for measurements}
\newacronym{rom}{ROM}{read-only memory}
\newacronym{ram}{RAM}{random-access memory}
\newacronym{cpu}{CPU}{central processing unit}
\newacronym{nvram}{NVRAM}{non-volatile random-access memory}
\newacronym{dram}{DRAM}{dynamic random-access memory}
\newacronym{ek}{EK}{endorsement key}
\newacronym{aik}{AIK}{attestation key}
\newacronym{ca}{CA}{certificate authority}
\newacronym{tpm}{TPM}{trusted platform module}
\newacronym{dtpm}{dTPM}{discrete TPM chip}
\newacronym{ftpm}{fTPM}{firmware TPM}
\newacronym{vtpm}{vTPM}{virtual TPM}
\newacronym{itpm}{iTPM}{integrated TPM}
\newacronym{pch}{PCH}{platform controller hub}
\newacronym{tcg}{TCG}{Trusted Computer Group}
\newacronym{pcr}{PCR}{platform configuration register}
\newacronym{pcrd}{dynamic PCR}{dynamic PCR}
\newacronym{pcrs}{static PCR}{static PCR}
\newacronym{ptt}{PTT}{Intel platform trusted technology}
\newacronym{uefi}{UEFI}{unified extensible firmware interface}
\newacronym{bios}{BIOS}{basic input/output system}
\newacronym{pxe}{PXE}{preboot execution environment}
\newacronym{svm}{SVM}{Secure Virtual Machine}
\newacronym{ima}{IMA}{integrity measurement architecture}
\newacronym{vpn}{VPN}{virtual private network}
\newacronym{daa}{DAA}{direct anonymous attestation}
\newacronym{loc}{LOC}{lines of code}
\newacronym{sloc}{SLOC}{source lines of code}
\newacronym{mee}{MEE}{memory encryption engine}
\newacronym{ias}{IAS}{Intel attestation service}
\newacronym{mrenclave}{MRENCLAVE}{enclave hash measurement}
\newacronym{acs}{IBM ACS}{IBM TPM attestation client-server}
\newacronym{cit}{Intel CIT}{Intel open cloud integrity technology}
\newacronym{initramfs}{initramfs}{initramfs}
\newacronym{vm}{VM}{virtual machine}
\newacronym{iaas}{IaaS}{Infrastructure-as-a-Service}
\newacronym{maas}{MaaS}{Metal-as-a-Service}
\newacronym{lpc}{LPC}{low pin count}
\newacronym{me}{CSME}{Intel converged security and manageability engine}
\newacronym{toctou}{TOCTOU}{time of check to time of use}
\newacronym{itl}{ITL}{Invisible Things Lab}
\newacronym{smm}{SMM}{system management mode}
\newacronym{dma}{DMA}{direct memory access}
\newacronym{tcb}{TCB}{trusted computing base}
\newacronym{bmc}{BMC}{baseboard management controller}
\newacronym{ipmi}{IPMI}{intelligent platform management interface}
\newacronym{nic}{NIC}{network interface card}
\newacronym{ssh}{SSH}{secure shell}
\newacronym{hsm}{HSM}{hardware security module}
\newacronym{vmm}{VMM}{virtual machine monitor}
\newacronym{kvm}{KVM}{kernel-based virtual machine}
\newacronym{qemu}{QEMU}{quick emulator}
\newacronym{mktme}{MKTME}{Intel multi-key total memory encryption}
\newacronym{tdx}{TDX}{Intel trust domain extensions}
\newacronym{tme}{TME}{Total Memory Encryption}
\newacronym{isp}{ISP}{internet service provider}
\newacronym{ssl}{SSL}{secure sockets layer}
\newacronym{ecc}{ECC}{elliptic-curve cryptography}
\newacronym{rsa}{RSA}{Rivest Shamir Adleman}
\newacronym{vmbr}{VMBR}{virtual-machine based rootkit}
\newacronym{aes}{AES}{advanced encryption standard}
\newacronym{sev}{SEV}{AMD secure encrypted virtualization}
\newacronym{dns}{DNS}{domain name system}
\newacronym{mitm}{MitM}{man-in-the-middle}
\newacronym{arp}{ARP}{address resolution protocol}
\newacronym{tcp}{TCP}{transmission control protocol}
\newacronym{mc}{MC}{monotonic counter}
\newacronym{mcs}{MCS}{monotonic counter service}
\newacronym{rest}{REST}{representational state transfer}
\newacronym{api}{API}{application programming interface}
\newacronym{cve}{CVE}{common vulnerabilities and exposures}
\newacronym{sriov}{SR-IOV}{single root input/output virtualization}
\newacronym{vtd}{VT-d}{Intel virtualization technology for directed I/O}
\newacronym{ecdsa}{ECDSA}{elliptic curve digital signature algorithm}
\newacronym{pal}{PAL}{piece of application logic}
\newacronym{gpu}{GPU}{graphical processing unit}
\newacronym{ml}{ML}{machine learning}
\newacronym{iommu}{IOMMU}{input-output memory management unit}
\newacronym{tcc}{TC}{trusted computing technologies}
\newacronym{ip}{IP}{intellectual property}
\newacronym{luks}{LUKS}{Linux unified key setup}
\newacronym{ai}{AI}{artificial intelligence}
\newacronym{gdpr}{GDPR}{general data protection regulation}
\newacronym{eu}{EU}{European Union}
\newacronym{dnn}{DNN}{deep neural networks}
\newacronym{tpu}{TPU}{tensor processing unit}
\newacronym{mpc}{MPC}{multi-party computation}
\newacronym{cnn}{CNN}{convolutional neural network}

\glsdisablehyper
\makeglossaries

\newcommand{\sysnospace}{{\rm\textsc{Perun}}}
\newcommand{\sys}{\sysnospace\xspace}
\newcommand{\sysIMA}{\sysnospace+IMA\xspace}
\newcommand{\sysIMASGX}{\sysnospace+IMA+SGX\xspace}
\newcommand{\sysIMAGPU}{\sysnospace+IMA+GPU\xspace}

\newcommand{\captionvspacesizetop}{0mm}
\newcommand{\captionvspacesizebottom}{3mm}

\newboolean{showcomments}
\setboolean{showcomments}{true}
\ifthenelse{\boolean{showcomments}}
{ \newcommand{\mynote}[3]{
    \fbox{\bfseries\sffamily\footnotesize#1}
    {\small$\blacktriangleright$\textsf{\emph{\color{#3}{#2}}}$\;\blacktriangleleft$}}}
    { \newcommand{\mynote}[3]{}}

\begin{document}

\title{\sys: Secure Multi-Stakeholder Machine Learning Framework with GPU Support}
\titlerunning{\sys}

\author{Wojciech Ozga\inst{1,3}\orcidID{0000-0002-0561-1279} \and
Do Le Quoc\inst{1,2}\orcidID{0000-0002-1433-0217} \and
Christof Fetzer\inst{1,2}\orcidID{0000-0001-8240-5420}}
\authorrunning{Ozga et al.}
\institute{TU Dresden, Germany \and Scontain UG, Germany \and  IBM Research - Zurich, Switzerland}

\maketitle

\begin{abstract}
\label{sec:abstract}

Confidential multi-stakeholder \gls{ml} allows multiple parties to perform collaborative data analytics while not revealing their \glsdesc{ip}, such as \gls{ml} source code, model, or datasets. 
State-of-the-art solutions based on homomorphic encryption incur a large performance overhead. 
Hardware-based solutions, such as \glspl{tee}, significantly improve the performance in inference computations but still suffer from low performance in training computations, \eg, deep neural networks model training, because of limited availability of protected memory and lack of GPU support.

To address this problem, we designed and implemented \sys, a framework for confidential multi-stakeholder machine learning that allows users to make a trade-off between security and performance. 
\sys executes ML training on hardware accelerators (\eg, GPU) while providing security guarantees using trusted computing technologies, such as \glsdesc{tpm} and \glsdesc{ima}.
Less compute-intensive workloads, such as inference, execute only inside TEE, thus at a lower \glsdesc{tcb}.
The evaluation shows that during the ML training on CIFAR-10 and real-world medical datasets, \sys achieved a $161\times$ to $1560\times$ speedup compared to a pure TEE-based approach. 


\glsresetall
\end{abstract}

\section{Introduction}
\label{sec:intro}

\begin{figure}[t]
	\centering
	\includegraphics[width=0.9\textwidth]{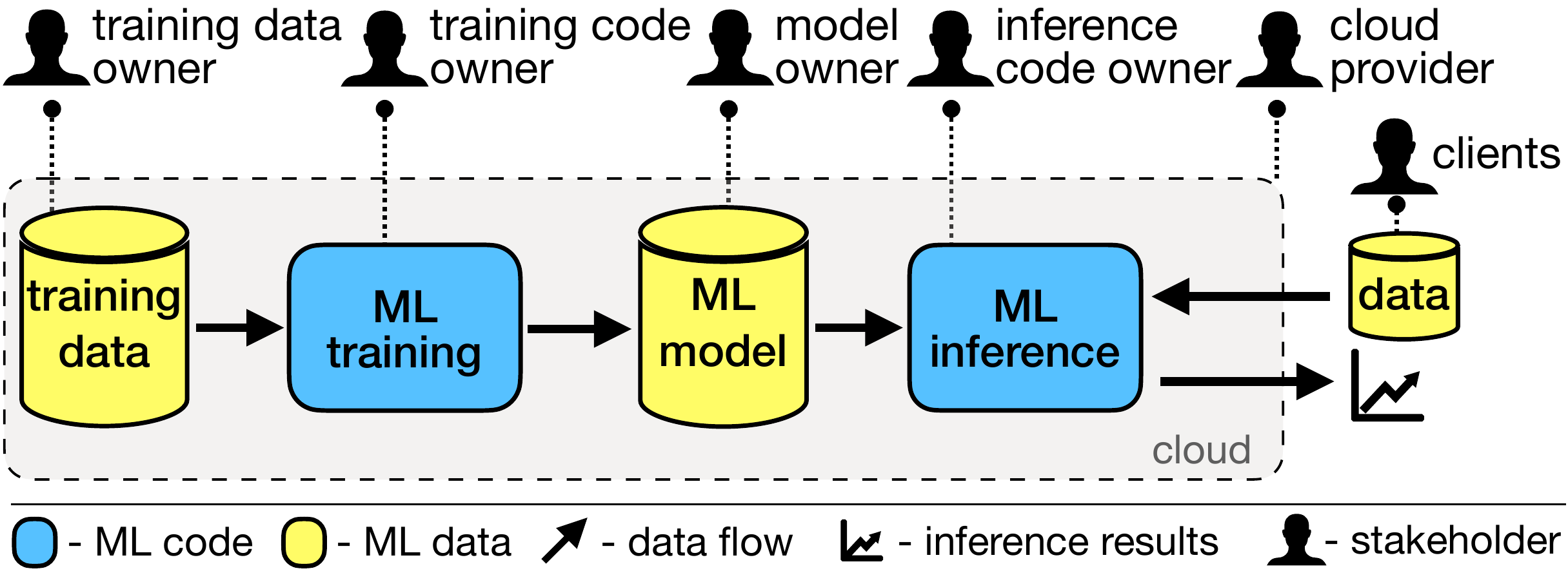}
	\vspace{\captionvspacesizetop}
	\caption{
		Stakeholders share source code, data, and computing power to build a ML application.
		They need a framework to establish mutual trust and share code and data securely.
	}
	\vspace{\captionvspacesizebottom}
	\label{fig:multistakeholder}
\end{figure}

\Gls{ml} techniques are widely adopted to build functional \gls{ai} systems. For example, face recognition systems allow paying at supermarkets without typing passwords; 
natural language processing systems allow translating information boards in foreign countries using smartphones; 
medical expert systems help to detect diseases at an early stage; image recognition systems help autonomous cars to identify road trajectory and traffic hazards. 
To build such systems, multiple parties or stakeholders with domain knowledge from various science and technology fields must cooperate since machine learning is fundamentally a multi-stakeholder computation, as shown in \autoref{fig:multistakeholder}. They would benefit from sharing their \gls{ip}--private training data, source code, and models--to jointly perform machine learning computations, to have high-accuracy machine learning models, only if they can ensure it remains confidential.

{\bf Training data owner}. 
\Gls{ml} systems rely on training data to build \emph{inference models}. 
However, the data is frequently sensitive and cannot be easily shared between disjoint entities.
For example, healthcare data used for training diagnostic models contain privacy-sensitive patient information.
The strict data regulations, such as \gls{gdpr}~\cite{gdpr}, impose an obligation on secure data processing.
Specifically, the training data must be under the training data owner's control and must be protected while at rest, during transmission, and training computation. 

{\bf Training code owner}. 
The training code owner implements a training algorithm that trains an inference model over the training data.
The training code (\eg, Python code) typically contains an optimized training model architecture and tuned parameters that build the business value and the inference model quality.
Thus, the training code is considered as confidential as training data. 
The training requires high computing power, and, as such, it is economically justifiable to delegate its execution to the cloud.
However, in the cloud, users with administrative access can easily read the training service source code implemented in popular programming languages, such as Python.

{\bf Model owner}.  
The inference model, the heart of any inference service, is created by training the model with a large amount of training data. 
It requires extensive computing power and is time-consuming and expensive.
Thus, the model owner, a training code owner, or a third party that buys the model, must protect the model's confidentiality.
 The trained models may reveal the privacy of the training data~\cite{deeplearning-DP}. 
Several works~\cite{model-inversion-attacks,deeplearning-DP} demonstrated that extracted images from a face recognition system look suspiciously similar to images from the underlying training data.

{\bf Inference code owner}. 
The inference code is an AI service allowing clients to use the inference model on a business basis.
The inference code is frequently developed using Python or JavaScript and hosted in the cloud.
Thus, the confidentiality of the  code and integrity of the computation must be protected. 

{\bf Inference data owner}.
The inference data owner, a client of an AI service, wants to protect his input data. 
For example, a person sends to a diagnostic service an X-ray scan of her brain to check for a brain tumor.
The inference data, \eg, a brain's scan, is privacy-sensitive and must not be accessible by the AI service provider.

To build an AI service, stakeholders must trust that others follow the rules protecting each other's IP.
However, it is difficult to establish trust among them.
First, some stakeholders might collude to gain advantages over the others~\cite{noor2013trust}.
Second, even a trustworthy stakeholder might lack expertise in protecting their IP from a skilled attacker gaining access to its computing resources~\cite{win8tradesecret,amazontradesecret}.
We tackle the following problem: \emph{How to allow stakeholders to jointly perform machine learning to unlock all AI benefits without revealing their IP?}

Recent works~\cite{delphi,cryptflow} demonstrated that cryptographic techniques, such as secure \glsdesc{mpc}~\cite{mpc1} and fully homomorphic encryption~\cite{fully-homomorphic}, introduce a large performance overhead, preventing their adoption for computing-intensive ML. 
Further works~\cite{ohrimenko,securetf} adopted \glspl{tee}~\cite{McKeen2013} to build ML systems showing that they are orders of magnitude faster than pure cryptographic solutions. 
TEE technologies provide a hardware-protected memory region called an \emph{enclave}. 
The CPU protects the integrity and confidentiality of applications executing inside an enclave against the operating system, the hypervisor, and the system administrator. 
Although promising for the ML inference, TEEs still incur significant overhead for memory-intensive computations such as the ML training because of the limited memory accessible to the enclave. 
Also, TEEs do not support hardware accelerators, \eg, \gls{gpu}.
TEEs are not enough for training processes, such as deep learning, that are computing- and memory-intensive because TEEs provide security guarantees only to applications executing on CPU.
We address the problem of: \emph{What trade-off between security and performance has to be made to allow the ML training to access hardware accelerators?}

We propose \sys, a framework allowing stakeholders to share their code and data only with certain ML applications running inside an enclave and on a trusted OS. 
\sys relies on encryption to protect the IP and on a trusted key management service to generate and distribute the cryptographic keys.
TEE provides confidentiality and integrity guarantees to ML applications and the key management service.
\Glsdesc{tcc}~\cite{trusted_computing_2009} provide integrity guarantees to the OS, allowing ML computations to access hardware accelerators.
Our evaluation shows that \sys achieves $0.96\times$ of native performance execution on the GPU and a speedup of up to $1560\times$ in training a real-world medical dataset compared to a pure TEE-based approach~\cite{securetf}.

Altogether, we make the following contributions: 
\setlist{nolistsep}
\begin{itemize}[noitemsep, leftmargin=4mm]
    \item We designed a secure multi-stakeholder ML framework that:
\emph{(i)} allows stakeholders to cooperate while protecting their IP (\S\ref{sec:highlevel}, \S\ref{sec:secretsharing})),
\emph{(ii)} allows stakeholders to select trade-off between the security and performance, allowing for hardware accelerators usage (\S\ref{sec:policy},\S\ref{sec:gpusupport}).
	\item We implemented \sys prototype (\S\ref{sec:implementation}) and evaluated it using real-world datasets (\S\ref{sec:evaluation}).
\end{itemize}

\section{Threat model}
\label{sec:threatmodel}

Stakeholders are financially motivated businesses that cooperate to perform ML computation.
Each stakeholder delivers an input (\eg, input training data, code, and ML models) as its \gls{ip}  for ML computations. 
The \gls{ip} must remain confidential during ML computations.
The stakeholders have limited trust. 
They do not share their \gls{ip} directly but encrypted so that only other stakeholders' applications, which source code they can inspect under a non-disclosure agreement or execute in a sandbox, can access the encryption key to decrypt it. 

An adversary wants to steal a stakeholder's \gls{ip} when it resides on a computer executing ML computation.
Such a computer might be provisioned in the cloud or a stakeholder's data center.
In both cases, an adversary has no physical access to the computer. We expect that data center access is well controlled and limited to trusted entities.

An adversary might exploit an OS misconfiguration or use social engineering to connect to the OS remotely. 
We assume she might get administrative access to the OS executing ML computation. 
It allows her to execute privileged software to read an ML process's memory.

The CPU with its hardware features, hardware accelerators, and secure elements (\eg, TPM) are trusted. 
We exclude side-channel attacks, \eg, ~\cite{Kocher2018spectre,Meltdown}.
We rely on the soundness of the cryptographic primitives used within software and hardware components.

\section{Design}
\label{sec:design}

Our objective is to provide an architecture that:
\setlist{nolistsep}
\begin{itemize}[noitemsep, leftmargin=4mm]
\item supports multi-stakeholder \gls{ml} computation,
\item requires zero code changes to the existing ML code,
\item allows for a trade-off between security and performance,
\item uses hardware accelerators for computationally-intensive tasks.
\end{itemize}

\subsection{High-level overview}
\label{sec:highlevel}
\begin{figure}[t]
	\centering
	\includegraphics[width=0.9\textwidth]{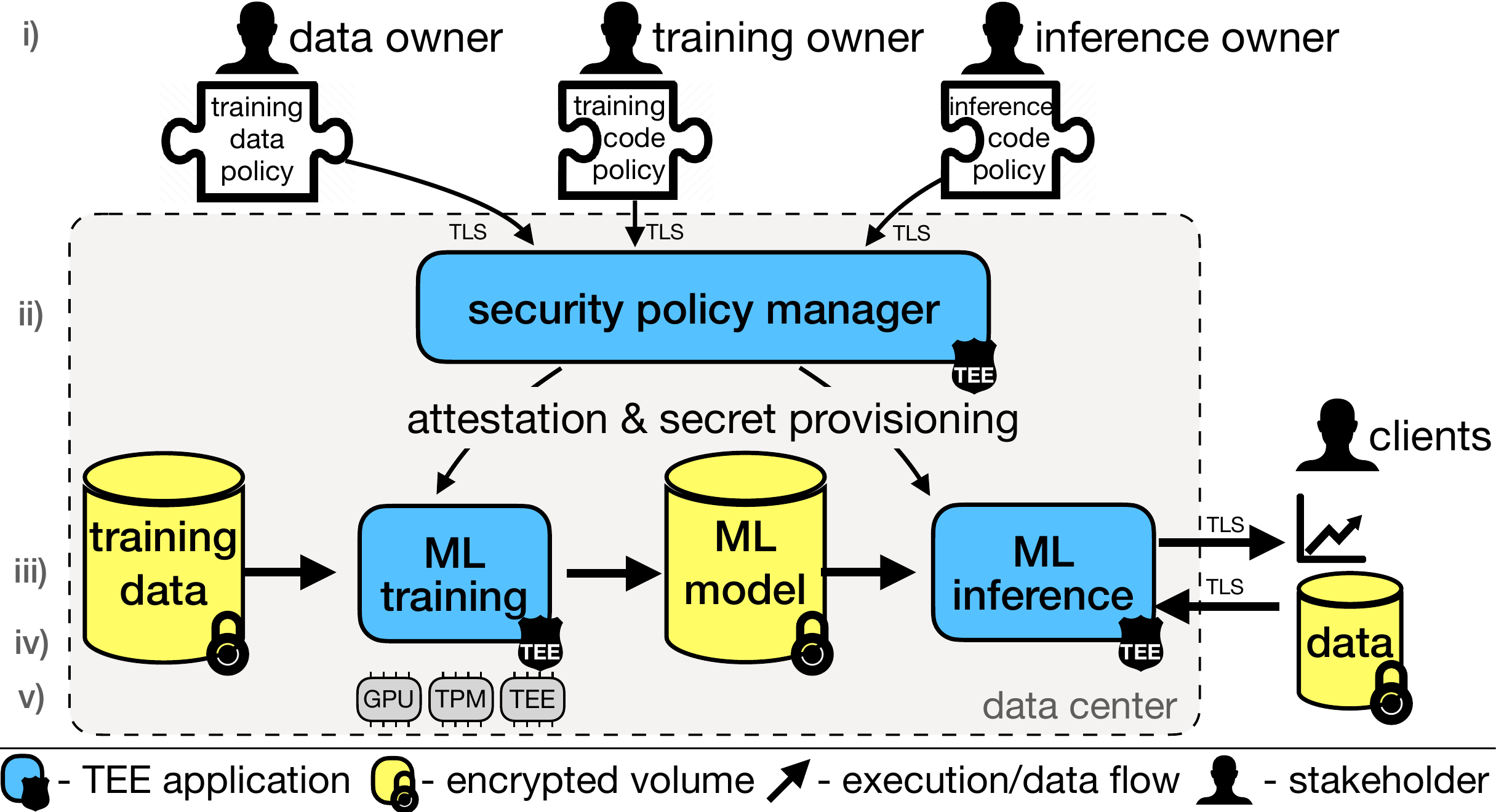}
	\vspace{\captionvspacesizetop}
	\caption{
		\sys framework supports multi-stakeholder \gls{ml} computation.
		Stakeholders trust the security policy manager.
		Inside security policies, they define which stakeholder's application can access a cryptographic key allowing decryption of confidential code or data.
		TEE protects code, data, and cryptographic keys.
	}
	\vspace{\captionvspacesizebottom}
	\label{fig:architecture}
\end{figure}

\autoref{fig:architecture} shows the \sys framework architecture that supports multi-stakeholder computation and the use of dedicated hardware accelerators.
The framework consists of five components:
(i) \emph{stakeholders}, the parties who want to perform ML jointly while keeping their IP protected;
(ii) \emph{security policy manager}, a key management and configuration service that allows stakeholders to share IPs for ML computations without revealing them;
(iii) \emph{ML computation} including training and inference; 
(iv) \emph{GPU}, hardware accelerators enabling high-performance ML computation; and
(v) \emph{TEE} and \emph{TPM}, hardware \emph{secure elements} enabling confidentiality and integrity of ML computations on untrusted computing resources, \eg, in the cloud.

To allow multiple stakeholders to perform ML and keep their IP confidential, we propose that the IP remains under the stakeholder's control. 
To realize that idea, we design the security policy manager that plays a role of the \emph{root of trust}. 
Stakeholders establish trust in this component using the \emph{remote attestation} mechanism, like \cite{johnson2016intel}), offered by a TEE. 
The TEE, \eg, \gls{sgx}~\cite{costan2016intel}, guarantees the confidentiality and integrity of processed code and data. 
After stakeholders ensure the security policy manager executes in the TEE, they submit to it security policies defining access control to their encryption keys.
Each stakeholder's IP is encrypted with a different key, and the security policy manager uses security policies to decide who can access which keys.
From a technical perspective, the security policy manager generates the keys inside the TEE and sends them only to ML computations executing inside the TEE. 
Thus, these keys cannot be seen by any human. 

Depending on individual stakeholders' security requirements, \sys offers different throughput/latency performances for ML computations. 
For stakeholders willing strong integrity and confidentiality guarantees, \sys executes ML computations only inside TEEs enclaves, \ie, input and output data, code, and models never leave the enclave.
On the other hand, for stakeholders accepting a larger trusted computing base in exchange for better performance, \sys enables \glsdesc{tcc}~\cite{trusted_computing_2009} to protect ML computations while executing them on hardware accelerators, \eg, \gls{gpu}.
\Gls{ima} is an integrity enforcement mechanism that prevents adversaries from running arbitrary software on the OS, \ie, software that allows reading data residing in the main memory or being transferred to or processed by the GPU. 
The security policy manager verifies that such a mechanism is enabled by querying a secure element compatible with the \gls{tpm}~\cite{tpm_2_0_spec} attached to the remote computer.

\subsection{Keys sharing}
\label{sec:secretsharing}

Stakeholders use security policies to share encryption keys protecting their IP. 
For example, the training data owner specifies in his security policy that he allows the ML computation of the training code owner to access his encryption key to decrypt the training data. 
The security policy manager plays a key role in the key sharing process.
It generates an encryption key inside the TEE and securely distributes it to ML computations accordingly to the security policy.
In the example above, the training code owner cannot see the shared secret because it is transferred to his application executing inside the TEE.

To provision ML computations with encryption keys, the security policy manager authenticates them using a remote attestation protocol offered by a TEE engine, such as the SGX remote attestation protocol~\cite{johnson2016intel}.
During the remote attestation, the TEE engine provides the security policy manager with a cryptographic measurement of the code executing on the remote platform.
The cryptographic measurement -- output of the cryptographic hash function over the code loaded by the TEE engine to the memory -- uniquely identifies the ML computation, allowing the security policy manager to authorize access to the encryption key based on the ML computation identity and stakeholder's security policies.

\subsection{Security policy}
\label{sec:policy}
\sys relies on security policies as a means to define dependencies among stakeholders computation and shared data.

\autoref{lst:policy} shows an example of a security policy. 
The policy has a unique name (line \ref{policy_name}), typically combining a stakeholder's name and its IP name.
The name is used among stakeholders to reference \emph{volumes} containing code, input, or output data.
A volume is a collection of files encrypted with an encryption key managed by the security policy manager.
Only authorized ML computations have access to the key required to decrypt the volume and access the IP.

The ML computation definition consists of a command required to execute the computation inside a container (line \ref{policy_exec}) and a cryptographic hash over the source code content implementing the ML computation (line \ref{policy_mrenclave}).
The security policy manager uses the hash to authenticate the ML computation before providing it with the encryption key.

The policy allows selecting trade-offs between security and performance. 
For example, a training code owner who wants to use the \gls{gpu} to speed up the ML training computation might define conditions under which he trusts the OS.
In such a case, a stakeholder defines a certificate chain permitting to verify the authenticity of a secure element attached to the computer (line \ref{policy_certificate}) and expected integrity measurements of the OS (lines \ref{policy_pcrs_start}-\ref{policy_pcrs_end}).
The security policy manager only provisions the ML computations with the encryption key if the OS integrity, \ie, kernel sources and configuration, are trusted by the stakeholder.
Specifically, the OS integrity measurements reflect what kernel code is running and whether it has enabled required security mechanisms.
Only then, the ML computations can access the confidential data and send it to the outside of the TEE, \eg, a GPU.

The stakeholder's public key is embedded inside the policy (line \ref{policy_creator}). 
The security policy manager accepts only policies containing a valid signature issued with a corresponding private key owned by a stakeholder.

\lstdefinestyle{interfaces}{float=tbp, floatplacement=tbp}
\begin{lstlisting}[caption=Security policy example,label={lst:policy},language=yaml,breaklines=true,breakatwhitespace=true,style=interfaces,escapechar=^]
name: training_owner/training_code ^\label{policy_name}^
command: python /app/training.py ^\label{policy_exec}^
volumes: ^\label{policy_volume_start}^
  - path: /training_data
    import: training_data_owner/training_data_service ^\label{policy_volumes_import}^
  - path: /inference_model
    export: inference_owner/inference_service ^\label{policy_volumes_export}^ ^\label{policy_volume_end}^
integrity_hash: {"0a11...bb3f"}  ^\label{policy_mrenclave}^
operating_system: ^\label{policy_tb_start}^
  certificate_chain: |- ^\label{policy_certificate}^
    -----BEGIN CERTIFICATE-----
    # certificate chain allowing 
    # verification of the secure
    # element manufacturer
    -----END CERTIFICATE-----
  integrity: ^\label{policy_pcrs_start}^
    measure0: e0f1...4be6 
    measure1: ae44...3a6e
    measure2: 3d45...796d ^\label{policy_pcrs_end}^ ^\label{policy_tb_end}^
stakeholder: |- 
  -----BEGIN PUBLIC KEY----- ^\label{policy_creator}^
  # the policy owner's public key
  -----END PUBLIC KEY-----
\end{lstlisting}

\subsection{Hardware ML accelerators support}
\label{sec:gpusupport}

Typically, ML computations (\eg, the \glsdesc{dnn} training) are extremely intensive because they must process a large amount of input data.
To decrease the computation time, popular ML frameworks, such as TensorFlow~\cite{tensorflow}, support hardware accelerators, such as GPUs or Google \glspl{tpu}.
Unfortunately, existing hardware accelerators do not support confidential computing, thus do not offer enough security guarantees to be used in the multi-stakeholder ML computation.
For example, an adversary who exploits an OS misconfiguration~\cite{misconfigurations_2013} can launch arbitrary software to read data transferred to the GPU from any process executing in the OS.
Even if ML computations execute inside the TEE enclaves, an adversary controlling the OS can read the data when it leaves the TEE, \ie, it is transferred to the GPU or is processed by the GPU.
Because of this, we design \sys to support additional security mechanisms protecting access to the data (also code and ML models) while being processed out of the TEE.
This also allows stakeholders to make a trade-off between security level and performance they want to achieve when performing ML computations.

\begin{figure}[tbp!]
\centering
    \includegraphics[width=0.9\textwidth]{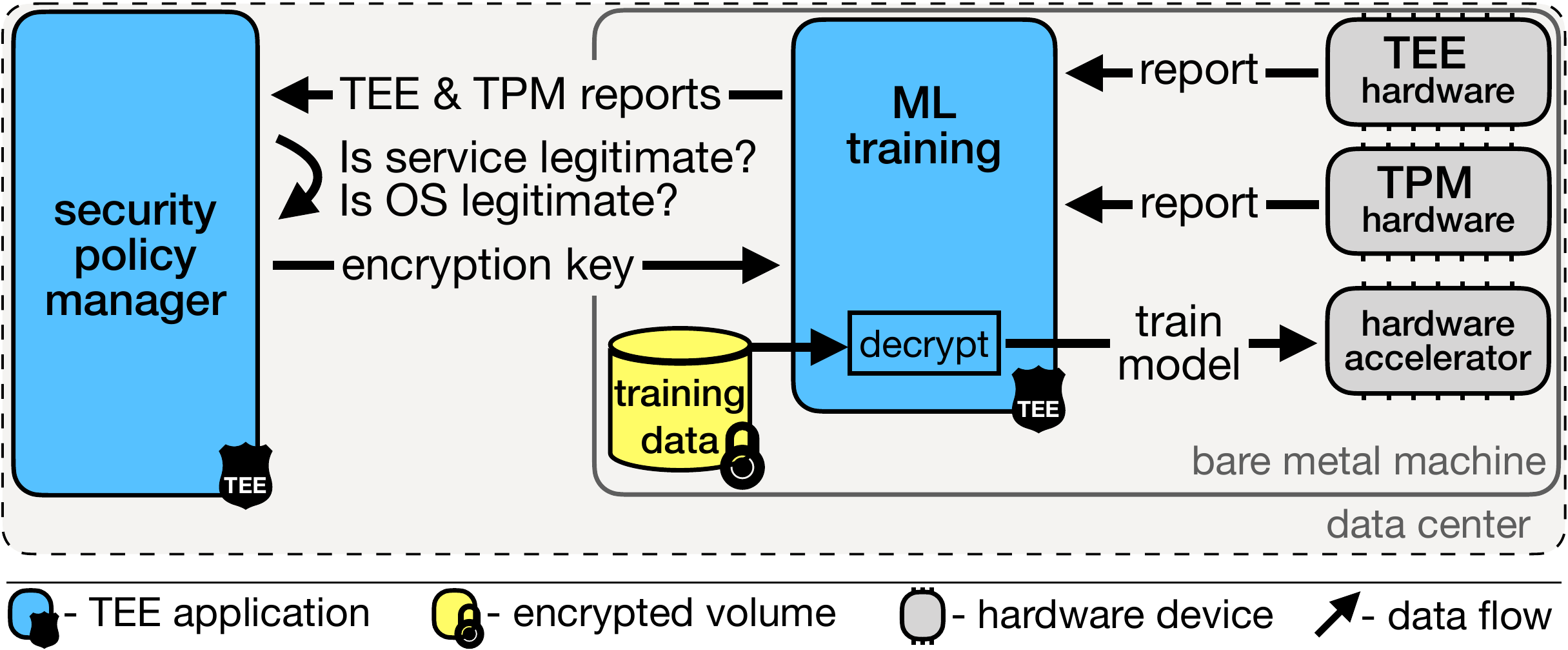}
    \vspace{\captionvspacesizetop}
    \caption{
        The high-level overview of \sys supporting secure computation using hardware accelerator, \eg, the GPU. 
        \sys performs both the SGX and TPM attestation before provisioning the ML code with cryptographic keys.
        The successful TPM attestation informs that the legitimate OS with enabled integrity-enforcement mechanisms controls access to the \gls{gpu}.
    }
    \vspace{\captionvspacesizebottom}
\label{fig:design}
\end{figure}

\autoref{fig:design} shows how \sys enables hardware accelerator support.
ML computations transfer to the security policy manager a report describing the OS's integrity state.
The report is generated and cryptographically signed by a secure element, \eg, a \gls{tpm} chip, physically attached to the computer.
The security policy manager authorizes the ML computation to use the encryption key only if the report states that the OS is configured with the required security mechanisms.
Specifically, the integrity enforcement mechanism, called \gls{ima}~\cite{ima_design_2004}, controls that the OS executes only software digitally signed by a stakeholder.
Even if an adversary gains root access to the system, she cannot launch arbitrary software that allows her to sniff on the communication between the ML computation and the GPU, read the data from the main memory, or reconfigure the system to disable security mechanisms.
This also allows \sys to mitigate a wide-range of side-channel attacks~\cite{Foreshadow,chen2018sgxpectre,weisse2018foreshadowNG,gotzfried_cache_2017} which are vulnerabilities of TEEs. 

To enable hardware accelerator support, a stakeholder specifies expected OS integrity measurements inside the security policy (\autoref{lst:policy}, lines \ref{policy_tb_start}-\ref{policy_tb_end}) and certificates allowing verification of the secure element identity.
The OS integrity measurements are cryptographic hashes over the OS's kernel loaded to the memory during the boot process.
A secure element collects such measurements during the boot process and certifies them using a private key linked to a certificate issued by its manufacturer. 
The certificate and integrity measurements are enough for the security policy manager to verify that the \gls{ima} enforces the OS integrity.

Although the hardware accelerator support comes at the cost of weaker security guarantees (additional hardware and software must be trusted compared to a pure TEE-based approach), it greatly improves the ML training computation's performance (see \S\ref{sec:eval:tradeoff}).

\subsection{Zero code changes}
\sys framework requires zero code changes to run existing ML computations, thus providing a practical solution for legacy ML systems.
To achieve it, \sys adapts platforms supporting running legacy applications inside the TEE, such as SCONE~\cite{arnautov2016scone} or Graphene\-SGX~\cite{tsai_graphene-sgx:2017}.
These platforms allow executing unmodified code inside the TEE by recompiling the code using dedicated cross-compilers or by running them with a modified interpreter executing in the TEE.

\subsection{Policy deployment and updates}
A stakeholder establishes a \gls{tls} connection to the security policy manager to deploy a policy.
During the TLS handshake, the stakeholder verifies the identity of the security policy manager.
The security policy manager owns a private key and corresponding certificate signed by an entity trusted by a stakeholder.
For example, such a certificate can be issued by a TEE provider who certifies that given software running inside a TEE and identified by a cryptographic hash is the security policy manager.
Some TEE engines, such as \gls{sgx}, offer such functionality preventing even a service administrator from seeing the private key~\cite{intel_sgxra_whitepaper}.
For other TEEs, a certificate might be issued by a cloud provider operating the security policy manager as part of cloud offerings.

\sys requires that the security policy manager authorizes changes to the deployed policy.
Otherwise, an adversary might modify the stakeholder's policy allowing malicious code to access the encryption key.
In the \sys design, the stakeholder includes his public key inside the digitally signed security policy.
Since then, the security policy manager accepts changes to the policy only if a new policy has a signature issued with the stakeholders' private key corresponding to the public key present in the existing policy.
By having a public key embedded in the security policy, other stakeholders can verify that the policy is owned by the stakeholder they cooperate with.

The details of the policy security manager regarding key management, high availability, tolerance, and protection against rollback attacks are provided in \cite{palaemon_2020}.

\section{Implementation}
\label{sec:implementation}
We implemented the \sys prototype based on TensorFlow version 2.2.0 and the SCONE platform~\cite{arnautov2016scone} since SCONE provides an ecosystem to run unmodified applications inside a TEE, including a key management system~\cite{scone_cas,palaemon_2020} to distribute the configuration to applications.
We rely on Intel SGX~\cite{costan2016intel} as a TEE engine because it is widely used in practice.

Our prototype uses a TPM chip~\cite{tpm_2_0_spec} to collect and report integrity measurements of the Linux kernel loaded to the memory during a trusted boot~\cite{drtm_tcg} provided by tboot~\cite{tboot} with \gls{txt}~\cite{intel_txt_whitepaper}. 
The Linux kernel is configured to enforce the integrity of software, dynamic libraries, and configuration files using Linux IMA~\cite{ima_design_2004}, a Linux kernel's security subsystem.
Using the TPM chip, \sys verifies that the kernel is correctly configured and stops working if requirements are not met.

We use an nvidia GPU as an accelerator for ML computation. The ML services are implemented in Python using TensorFlow framework, which supports delegating ML computation to the GPU. 

\subsection{Running ML computations inside Intel SGX}
To run unmodified ML computations inside the SGX enclaves, we use the SCONE cross-compiler and SCONE-enabled Python interpreters provided by Scontain as Docker images.
These images allow us to build binaries that execute inside the SGX enclave or run Python code inside SGX without any source code modification.

The SCONE wraps an application in a dynamically linked loader program (\emph{SCONE loader}) and links it with a modified C-library (\emph{SCONE runtime}) based on the musl libc~\cite{musl_libc}.
On the ML computation startup, the SCONE loader requests SGX to create an isolated execution environment (enclave), moves the ML computation code inside the enclave, and starts. 
The SCONE runtime, which executes inside the enclave along with the ML computations, provides a sanitized interface to the OS for transparent encryption and decryption of data entering and leaving the enclave. 
Also, the SCONE runtime provides the ML computations with its configuration using configuration and attestation service (CAS)~\cite{scone_cas}.

\subsection{Encryption key sharing}
We implement the security policy manager in the \sys architecture using the CAS, to generate, distribute, and share encryption keys between security policies. We decided to use the CAS because it integrates well with SCONE-enabled applications and implements the SGX attestation protocol~\cite{johnson2016intel}. Other key management systems supporting the SGX attestation protocol might be used \cite{chakrabarti2017intel,fortanix_kms} but require additional work to integrate them into SCONE. 

We create a separate CAS policy for each stakeholder.
The policy contains an identity of the stakeholder's IP (data, code, and models) and its access control and configuration.
It is uploaded to CAS via the mutual TLS authentication using a stakeholder-specific private key corresponding to the public key defined inside the policy. This fulfills the \sys requirement of protecting unauthorized stakeholders from modifying policies.
The IP identity is defined using a unique per application cryptographic hash calculated by the SGX engine over the application's pages and their access rights. 
The SCONE provides this value during the application build process.
The CAS allows for the specification of the encryption key as a program argument, environmental variable, or indirectly as a key related to an encrypted volume.
Importantly, the CAS allows defining which policies have access to the key. Thus, with the proper policy configuration, stakeholders share keys among enclaves as required in the \sys architecture.

Our prototype uses the CAS encrypted volume functionality, for which the SCONE runtime fetches from the CAS the ML computation configuration containing the encryption key.
Specifically, following the SGX attestation protocol, the SCONE runtime sends to CAS the SGX attestation report in which the SGX hardware certifies the ML computation identity.
The CAS then verifies that the report was issued by genuine SGX hardware and the ML computation is legitimate.
Only afterward, it sends to the SCONE runtime the encryption key. 
The SCONE runtime transparently encrypts and decrypts data written and read by the ML computation from and to the volume.
The ML computations, \ie, training and inference authorized by stakeholders via policies, can access the same encryption key, thus gaining access to a shared volume.

\subsection{Enabling GPU support with integrity enforcement}
Our prototype implementation supports delegating ML computations to the GPU under the condition that the communication between the enclave and the GPU is handled by the integrity-enforced OS. 
The integrity enforcement mechanism prevents intercepting confidential data leaving the enclave because it limits the OS functionality to a subset of programs essential to load the ML computation and the GPU driver.
Thus, a malicious program cannot run alongside the ML computations on the same computing resources.
We use trusted boot and TPM to verify it, \ie, that the remote computer runs a legitimate Linux kernel with enabled integrity enforcement that limits software running on the computer to the required OS services, the GPU driver, and ML computations.

\textbf{Trusted boot.}
\Glsdesc{tcc} define a set of technologies that measure, report, and enforce kernel integrity.
Specifically, during the computer boot, we rely on a trusted bootloader~\cite{tboot}, which uses a hardware CPU extension~\cite{intel_txt_whitepaper} to measure and securely load the Linux kernel to an isolated execution environment~\cite{drtm_tcg}.
The trusted bootloader measures the kernel integrity (a cryptographic hash over the kernel sources) and sends the TPM chip measurements.

The TPM stores integrity measurements in a dedicated tamper-resistant memory called \glspl{pcr}.
A PCR value cannot be set to an arbitrary value.
It can only be extended with a new value using a cryptographic hash function: \emph{PCR\_extend = hash(PCR\_old\_value $\vert\vert$ data\_to\_extend)}.
This prevents tampering with the measurements after they are extended to a PCR.
The TPM implements a TPM attestation protocol~\cite{tcg_tpm_attestation} in which it uses a private key known only to the TPM to sign a report containing PCRs.
In our prototype, we use the TPM attestation protocol to read the TPM report certifying that an ML computation executes on an integrity-enforced OS, \ie, a Linux kernel with enabled \gls{ima}.

\textbf{Integrity enforcement.}
\begin{figure}[tbp!]
    \centering
    \includegraphics[width=0.9\textwidth]{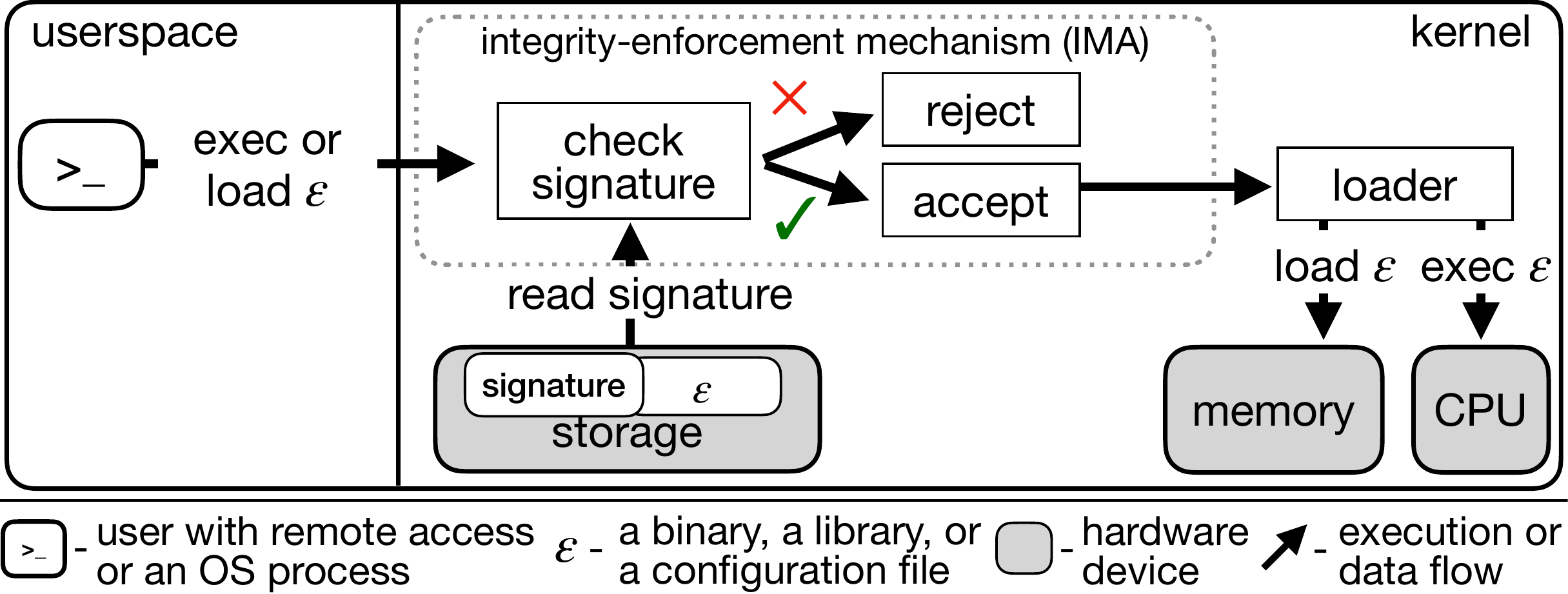}
    \vspace{\captionvspacesizetop}
    \caption{
        The kernel integrity-enforcement system authenticates a file by checking its digital signature before loading it to the memory.
    }
    \vspace{\captionvspacesizebottom}
    \label{fig:IMA}
\end{figure}
\gls{ima} is a kernel mechanism that authenticates files before allowing them to be loaded to the memory. 
\autoref{fig:IMA} shows how the \gls{ima} works.
A process executing in userspace requests the kernel to execute a new application, load a dynamic library, or read a configuration file. 
IMA calculates the cryptographic hash over the file's content, reads the file's signature from the file's extended attribute, and verifies the signature using a public key stored in the kernel's \emph{\.ima keyring}. 
If the signature is correct, IMA extends the hash to a dedicated PCR and allows the kernel to continue loading the file.

\textbf{Trusted boot service.}
Because SCONE is proprietary software, we could not modify the SCONE runtime to provide the CAS with the TPM report.
Instead, we implemented this functionality in a \emph{trusted boot service} that uses the TPM to verify that the ML computations execute in the integrity-enforced OS.

The CAS performs the SGX attestation of the trusted boot service and provisions it with the TPM certificate as well as a list of the kernel integrity measurements. The trusted boot service reads the integrity measurements stored in PCRs using the TPM attestation protocol. The TPM genuineness is ensured by verifying the TPM certificate using a certificate chain provided by the CAS. The Linux kernel integrity is verified by comparing the integrity measurements certified by the TPM with the measurements read from the CAS.

We implemented the trusted boot service as an additional stage in the ML data processing.
It enables other ML computations to access the confidential data only if the OS state conforms to the stakeholder's security policy. 
It copies the confidential data from an encrypted volume of one ML computation to a volume accessible to another ML computation after verifying the kernel integrity using the TPM. 
Our implementation is complementary with \gls{luks}~\cite{linux_luks}. LUKS ensures that the kernel can decrypt the file system, \ie, ML computations and the encrypted volume, only if the kernel integrity has not changed. 
This prevents accessing the trusted boot service's volume after modifying the kernel configuration, \ie, disabling the integrity-enforcement mechanism.

\section{Evaluation}
\label{sec:evaluation}
\label{sec:eval:tradeoff}

\textbf{Testbed.} 
Experiments were executed on a ASUS Z170-A mainboard equipped with an Intel Core i7-6700K CPU supporting SGXv1, 
Nvidia GeForce RTX 2080 Super,
64\,GiB of RAM, 
Samsung SSD 860 EVO 2\,TB hard drive, 
Infineon SLB 9665 TPM\,2.0, 
a 10\,Gb Ethernet network interface card connected to a 20\,Gb/s switched network. 
The hyper-threading is enabled. 
The \gls{epc} is configured to reserve 128\,MB of RAM.
CPUs are on the microcode patch level 0xe2.
We run Ubuntu 20.04 with Linux kernel 5.4.0-65-generic. 
Linux IMA is enabled. The hashes of all OS files are digitally signed using a 1024-bit RSA asymmetric key.
The signatures are stored inside files' extended attributes, and the certificate signed by the kernel's build signing key is loaded to the kernel's keyring during initrd execution.

\textbf{Datasets.}
We use two datasets: 
\emph{(i)} the classical CIFAR-10 image dataset~\cite{krizhevsky2009learning}, and 
\emph{(ii)} the real-world medical dataset~\cite{medical-dataset}.

\subsection{Attestation latency}
We run an experiment to measure the overhead of verifying the OS integrity using the TPM.
Specifically, we measure how much time it takes an application implementing the trusted boot service to receive configuration from the security policy manager, read the TPM, and verify the OS integrity measurements.

The security policy manager executes on a different machine located in the same data center. 
It performs the SGX attestation before delivering a configuration containing two encryption keys--typical setup for ML computations--and measurements required to verify the OS integrity.
The security policy manager and the trusted boot service execute inside SCONE-protected Docker containers.

\autoref{table:attestation} shows that launching the application inside a SCONE-protected container takes 1573\,ms.
Running the same application that additionally receives the configuration from the security policy manager incurs 118\,ms overhead.
Additional 719\,ms are required to read the TPM quote, verify the TPM integrity and authenticity, and compare the read integrity measurements with expected values provided by the security policy manager. 
As we show next, 2.5\,sec overhead required to perform SGX and TPM attestation is negligible considering the ML training execution time.

\begin{table}[tbp]
	\small
	\setlength{\tabcolsep}{3pt}
	\center
	\caption{End-to-end latency of verifying software authenticity and integrity using SGX and TPM attestation. Mean latencies are calculated as 10\% trimmed mean from ten independent runs. \emph{sd} stands for standard deviation.}
	\begin{tabular}{>{\arraybackslash}p{0.48\columnwidth}>{\arraybackslash}p{0.43\columnwidth}}
		\rowcolor{gray!25}
		\textbf{}  &  \textbf{Execution time}  \\
		\hline
		Application in a container & 1573\,ms (sd=16\,ms) \\		
		\hline
		\quad + SGX attestation & 1691\,ms (sd=37\,ms) \\		
		\hline
		\quad + SGX and TPM attestation & 2410\,ms (sd=33\,ms) \\
		\hline
	\end{tabular}
	\vspace{2mm}
	\label{table:attestation}
\end{table}

\subsection{Security and performance trade-off }

To demonstrate the advantage of \sys in allowing users to select the trade-off between security and performance, we compare the performance of different security levels provided by \sys and the pure SGX based system called SecureTF~\cite{securetf}. 
We run the model training using the setups:
\emph{i)} only CPU (\emph{Native});
\emph{ii)} GPU (\emph{Native GPU});
\emph{iii)} \sys, IMA enabled (\emph{\sysIMA});
\emph{iv)} \sys, IMA and SGX enabled (\emph{\sysIMASGX});
\emph{v)} \sys with GPU, IMA enabled (\emph{\sysIMAGPU}).
In all setups, the trusted boot service executes inside the enclave.

\textbf{CIFAR-10 dataset.} 
We perform training using the CIFAR-10 dataset, a \glsdesc{cnn} containing four {\em conv} layers followed by two fully-connected layers. 
We use \emph{BatchNorm} after each conv layer. 
We apply the \emph{ADAM} optimization algorithm~\cite{adam} with the learning rate set to 0.001.

\begin{figure*}
	\centering
	\includegraphics[width=0.9\textwidth]{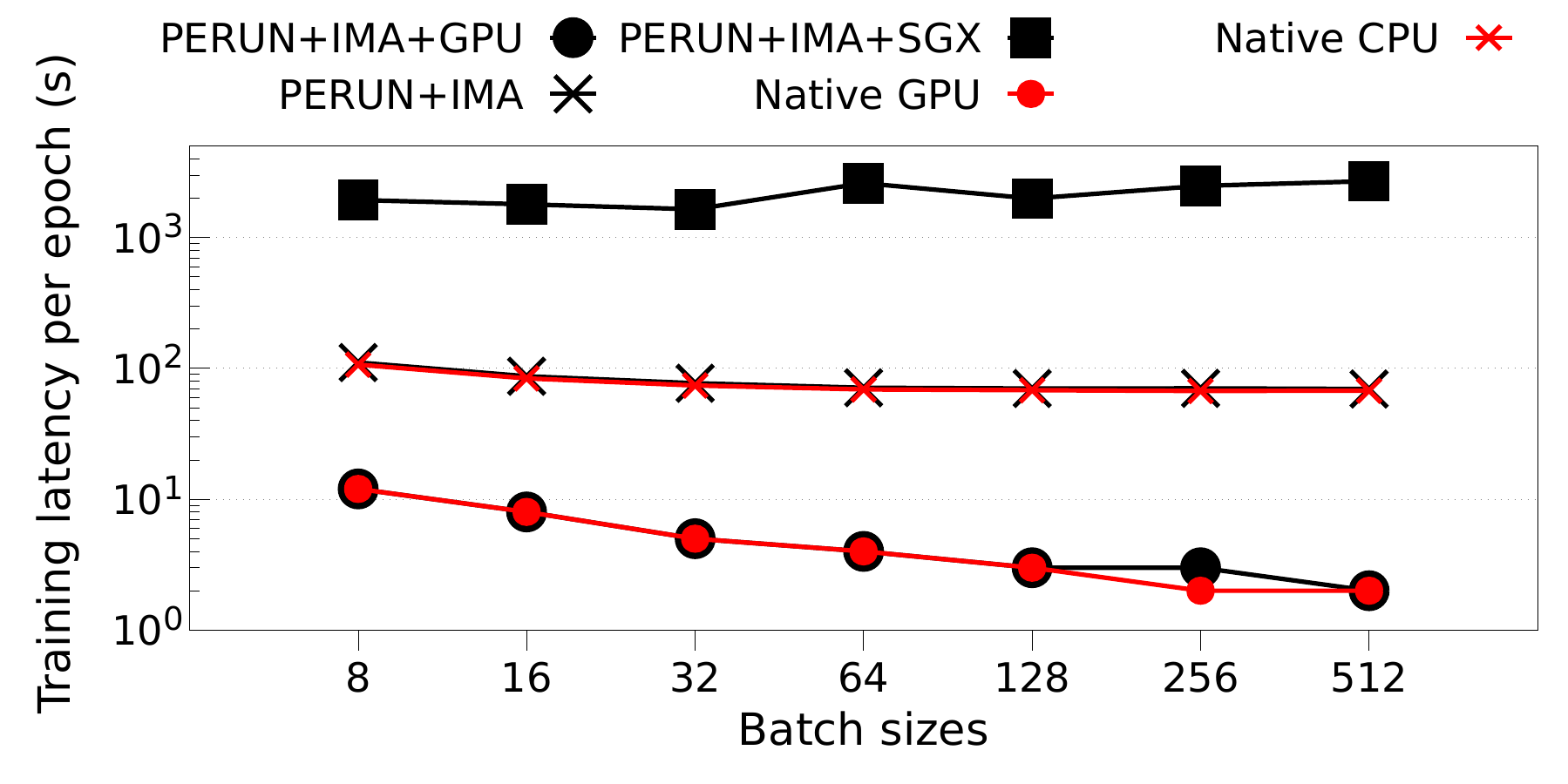}
	\caption{
		The CIFAR-10 training latency comparison among different security levels offered by \sys. 
		Mean latencies are calculated from five independent runs.
	}
	\label{fig:micro-cifar-latency}
\end{figure*}

\begin{figure*}
	\centering
	\includegraphics[width=0.9\textwidth]{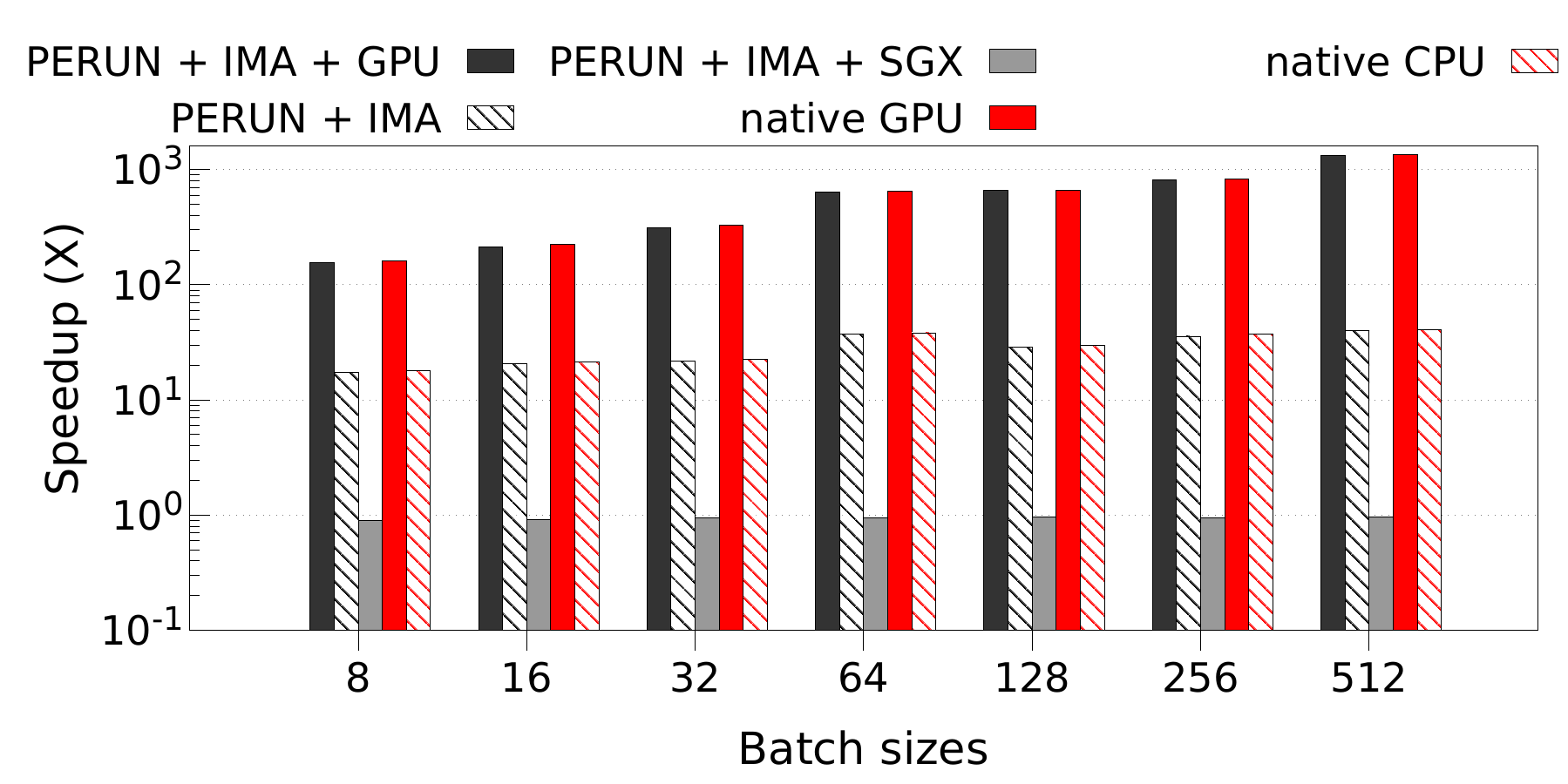}
	\caption{
		The CIFAR-10 training speedup of evaluated systems in comparison to \sys with the highest security level (\sys + IMA + SGX).
	}
	\label{fig:micro-cifar-speedup}
	\vspace{3mm}
\end{figure*}

\autoref{fig:micro-cifar-latency} shows the training latency  and \autoref{fig:micro-cifar-speedup} shows the \sys speedup depending on setups and batch sizes.
At the highest security level (\emph{\sysIMASGX}), \sys achieves almost the same performance as the pure SGX-based system, secureTF. 
This is because the training data is processed only inside the enclave and SGX performs compute-intensive paging caused by the limited \gls{epc} size (128\,MB) that cannot accommodate the training computation data (8\,GB).
When relying just on the integrity protection mechanism, \emph{\sysIMAGPU} and \emph{\sysIMA} achieve $1321\times$ and $40\times$ speedup compared to secureTF (batch size of $512$). 
With these setups, the \sys performance is similar to native systems ($\sim0.96\times$ of native latency) because the integrity protection mechanism performs integrity checks only when it loads files to the memory for the first time, leading to almost native execution afterward.

\textbf{Real-world medical dataset.} 
Next, we evaluate \sys using a large-scale real-world medical dataset~\cite{medical-dataset}. 
The dataset contains a wide range of medical images, including images of cancer and tumor treatment regimens for various parts of the human body, \eg, brain, colon, prostate, liver, and lung. 
It was created via CT or MRI scans by universities and research centers from all around the world. 
We perform training over the brain tumor images dataset (6.1\,GB) using the 2-D U-Net~\cite{unet} TensorFlow architecture from Intel AI~\cite{intel-unet}. 
It makes use of the \emph{ADAM} optimizer that includes 7\,760\,385 parameters with 32 features maps. 
We set the learning rate to 0.001 and the batch size to 32.

\autoref{table:medical-comp} shows that at the highest security level (the data is processed inside the enclave) {\emph{\sysIMASGX}} achieves the same performance as the referenced SGX-based system.
However, when relying just on the integrity protection mechanism to protect the data, \emph{\sysIMAGPU} and \emph{\sysIMA} achieve a speedup of $1559\times$ and $47\times$ compared to secureTF, respectively. 
We maintain the accuracy of 0.9875 in all experiments. 

\begin{table}[bp!]
	\small
	\setlength{\tabcolsep}{3pt}
	\center
	\vspace{4mm}
	\caption{The training latency comparison among different security levels of \sys, secureTF, and native. The results were obtained from a single run.}
	\begin{tabular}{>{\arraybackslash}p{0.44\columnwidth}>{\RaggedLeft\arraybackslash}p{0.32\columnwidth}>{\RaggedLeft\arraybackslash}p{0.15\columnwidth}}
		\rowcolor{gray!25}
		\textbf{System}  &  \textbf{Latency per epoch} & \textbf{Speedup}  \\
	    \hline
	    Native CPU & 5\,h 26\,min 14\,sec & $47\times$ \\
		\hline
		Native GPU & 9\,min 54\,sec & $1561\times$ \\
	    \hline		
		\sysIMA & 5\,h 26\,min 17\,sec & $47\times$ \\
		\hline
	    \sysIMASGX & 257\,h 27\,min 49\,sec & $\sim1\times$ \\
		\hline
		\sysIMAGPU & 9\,min 55\,sec & $1560\times$ \\
		\hline
	    secureTF & 257\,h 43\,min 53\,sec & (baseline) \\
	    \hline
	\end{tabular}
	\label{table:medical-comp}
\end{table}
\section{Related work}
\label{sec:related}

{\bf Secure multi-party computation.} 
Although cryptographic schemes, such as secure \gls{mpc} and fully homomorphic encryption, are promising to secure multi-stakeholder ML computation, they have limited application in practice~\cite{conclave,ohrimenko}.
They introduce high-performance overhead~\cite{ohrimenko,delphi,cryptflow,secureml,gazelle} (limiting factor for computing-intensive ML) and require to heavily modify existing ML code. 
Furthermore, they do not support all ML algorithms, such as, deep neural networks.
Some of them require additional assumptions, like \gls{mpc} protocol that requires a subset of stakeholders to be honest. 
Unlike \sys, most of them lack support for training computation.

{\bf Secure ML using TEEs.} 
Many works leverage TEE to support secure ML~\cite{privado,chiron,ohrimenko}. 
Chiron~\cite{chiron} uses SGX for privacy-preserving ML services, but it is only a single-threaded system. 
Also, it needs to add an interpreter and model compiler into the enclave.
This incurs high runtime overhead due to the limited \gls{epc} size. 
The work from Ohrimenko et al.~\cite{ohrimenko} also relies on SGX for secure ML computations. 
However, it does not allow using hardware accelerators and supports only a limited number of operators --- not enough for complex ML computations. 
In contrast to these systems, \sys supports legacy ML applications without changing their source code. 
SecureTF~\cite{securetf} is the most relevant work for \sys because it also uses SCONE.
It supports inference and training computation, as well as distributed settings. 
However, it is not clear how secureTF can be extended to support secure multi-stakeholders ML computation. 
Also, secureTF does not support hardware accelerators, making it less practical for training computation. 
Other works~\cite{slalom,goten,goat} use SGX and untrusted GPUs for secure ML computations. 
They split ML computations into trusted parts running in the enclave and untrusted parts running in the GPU. 
However, they require changing the existing code and do not support multi-stakeholder settings.

{\bf Trusted GPUs}.  
Although trusted computation on GPUs is not commercially available, there is ongoing research. 
HIX~\cite{hix} enables memory-mapped I/O access from applications running in SGX by extending an SGX-like design with duplicate versions of the enclave memory protection hardware. 
Graviton~\cite{graviton} proposes hardware extensions to provide TEE inside the GPU directly. Graviton requires to modify the GPU hardware to disable directly accessing to the critical GPU interfaces, e.g., page table and communication channels from the GPU driver. 
Telekine~\cite{telekine} restricts access to GPU page tables without trusting the kernel driver, and it secures communication with the GPU using cryptographic schemes. 
The main limitation of these solutions is that they require hardware modification of the GPU design, so they cannot protect existing ML computations and they also do not support multi-stakeholder ML computatitions.

\section{Conclusion}
\label{sec:conclusion}

\sys allows multiple stakeholders to perform \gls{ml} without revealing their intellectual property. 
It provides strong confidentiality and integrity guarantees at the performance of existing TEE-based systems. 
With the help of trusted computing, \sys permits utilizing hardware accelerators, reaching native hardware-accelerated systems' performance at the cost of a larger trusted computing base. 
When training an ML model using real-world datasets, \sys achieves $0.96\times$ of native performance execution on the GPU and a speedup of up to $1560\times$ compared to a state-of-the-art SGX-based system.

\pagebreak
\bibliographystyle{splncs04}
\interlinepenalty=10000
\bibliography{reference}

\end{document}